\documentclass{article}

     \PassOptionsToPackage{numbers, compress}{natbib}

\usepackage[final]{cml4impact_2022}

\usepackage{natbib} 
\usepackage[utf8]{inputenc} 
\usepackage[T1]{fontenc}    
\usepackage{hyperref}       
\usepackage{url}            
\usepackage{booktabs}       
\usepackage{amsfonts}       
\usepackage{nicefrac}       
\usepackage{microtype}      
\usepackage{xcolor}         

\usepackage{graphicx}
\usepackage{overpic}
\usepackage{comment}
\usepackage{array}
\usepackage{amsmath}

\usepackage{caption}
\usepackage{subcaption}

\newcommand{\Real}{\mathbb{R}}
\newcommand{\indep}{\perp \!\!\! \perp}

\DeclareMathOperator*{\argmax}{arg\,max}

\title{Identifying the Causes of Pyrocumulonimbus (PyroCb)}

\author{
        Emiliano Díaz Salas-Porras\\
        Image Processing Laboratory\\
        University of Valencia\\
        Valencia, Spain\\
        \texttt{emiliano.diaz@uv.es}\\
        \And
        Kenza Tazi\\ 
        Department of Engineering\\
        University of Cambridge\\
        Cambridge, UK\\
        \texttt{kt484@cam.ac.uk}\\
        \And
        Ashwin Braude\\
        Laboratoire Atmosphères,\\
        Milieux, Observations Spatiales\\
        Institut Pierre-Simon Laplace\\
        Guyancourt, France\\
        \And
        Daniel Okoh\\
        Centre for Atmospheric Research\\
        National Space Research \\
        and Development Agency\\
        Abuja, Nigeria\\
        \And
        Kara D. Lamb\\
        Department of Earth and \\
        Environmental Engineering\\
        Columbia University\\
        New York, US\\
        \And
        Duncan Watson-Parris\\
        Department of Atmospheric,\\ 
        Oceanic and Planetary Physics\\
        University of Oxford\\
        Oxford, UK\\
        \And
        Paula Harder\\
        Fraunhofer Institute for \\
        Industrial Mathematics\\
        Kaiserslautern, Germany\\
        \And
        Nis Meinert\\
        Pasteur Labs \\
        New York, US\\
        }

\begin{document}

\maketitle

\begin{abstract}
  A first causal discovery analysis from observational data of pyroCb (storm clouds generated from extreme wildfires) is presented. Invariant Causal Prediction was used to develop tools to understand the causal drivers of pyroCb formation. This includes a conditional independence test for testing $Y \indep E|X$ for binary variable $Y$ and multivariate, continuous variables $X$ and $E$, and a greedy-ICP search algorithm that relies on fewer conditional independence tests to obtain a smaller more manageable set of causal predictors. With these tools, we identified a subset of seven causal predictors which are plausible when contrasted with domain knowledge: surface sensible heat flux, relative humidity at $850$\,hPa, a component of wind at $250$\,hPa, $13.3$\,\textmu m thermal emissions, convective available potential energy, and altitude.
\end{abstract}

\section{Introduction}

More than $17$\,million people have been negatively affected and $\$144$ billion USD lost during major wildfire events occurring during the last $30$ years \citep{guha2022}. A changing climate further increases the risk that wildfires pose to people and the environment. By the end of the century, the frequency of wildfires is expected to increase by a factor of $1.3$ to $1.6$, compared to a 2000-2010 reference period \citep{unep2022}. The number of extreme wildfires is increasing to an even greater degree \citep{dowdy2019future}. Furthermore, the degradation of air quality due to the creation of aerosols and ozone from fires resulted in between $260\,000$ and $600\,000$ premature deaths each year \citep{johnston2012estimated}.

Pyrocumulonimbus (pyroCbs) are storm clouds generated by large and intense wildfires \citep{peterson2017detection, rodriguez2020extreme}. These storm clouds generate their own weather fronts which can make wildfire behaviour unpredictable: strong winds fan the flames of the fire and change its direction while lightning ignites new fires in the vicinity \citep{peterson2021australia}. This unpredictability increases dangers for firefighters and other first responders, and reduces timescales for evacuating surrounding communities. PyroCbs also transport wildfire aerosols into the stratosphere where they may remain for several months \citep{peterson2021australia, yu2019black, peterson2018wildfire}. These events, which can transport amounts of aerosols on the scale of volcanic eruptions, may have important impacts on the Earth's climate, e.g., through direct absorption of solar radiation or cloud formation \citep{ke2021global, stocker2021observing, kablick2020australian}. PyroCbs may also disrupt the recovery of the ozone layer \citep{rieger2021stratospheric, schwartz2020australian}.

Despite the risk posed by pyroCbs, the conditions leading to their occurrence and evolution are still poorly understood \citep{peterson2018model,tory2018thermodynamics,tory2021pyrocumulonimbus}. Although it is well-known that pyroCbs are generated solely from large and intense wildfires, surface-based variables and fire weather indices (such as the Haines index \citep{haines1988}) have been shown to be poor predictors of the potential for pyroCb formation \citep{peterson2015}. The meteorological environment in which intense fires develop is also an important precursor of pyroCb formation. As smoke plumes rise in the atmosphere, they cool and entrain surrounding colder, dryer air, potentially reducing condensation; it is still not clear how significant the role played by aerosol particles, moisture, and heat released by the fire itself is to pyroCb formation \cite{tory2018thermodynamics,tory2021pyrocumulonimbus}. Several competing factors play a role, as the boundary layer must be dry enough to support intense fire activity and wildfire spread rates, while also supporting conditions favorable for thunderstorm formation. Recent studies suggest that both a deep, dry, and unstable lower troposphere, combined with a moist and unstable mid-troposphere are key ingredients for producing favorable conditions for pyroCb \cite{peterson2018model}. 

This study presents an analysis of the causes of pyroCb and is part of the \textsc{Pyrocast} machine learning pipeline that was developed with the aim of detecting, forecasting and understanding the drivers of pyroCb events. This pipeline was developed by the authors of the study presented here and is published in a companion paper \citep{pyrocast}. This study is structured as follows: in Section~\ref{pyrocastDB}, the \textsc{Pyrocast} database is summarised; the Invariant Causal Prediction methodology is introduced in Section~\ref{ICP};  implementation details are described in Section~\ref{setup} and, finally, results are  discussed in Section~\ref{section:results}.

The \textsc{Pyrocast} database, all code used in its construction, and all code used to produce results in this work are publicly available at \url{https://spaceml.org/} (\url{https://doi.org/10.56272/fpib2524}).

\section{\textsc{Pyrocast} Database}
\label{pyrocastDB}
The \textsc{Pyrocast} database contains geostationary satellite images, meteorological, and fuel data associated with individual wildfire events. It also contains pixel-wise labels indicating whether the wildfire produced a pyroCb at given space-time coordinates.
The \textsc{Pyrocast} database contains four sources of information: ($1$) images from geostationary satellites: six channels; ($2$) meteorological information: $15$ variables; ($3$) fuel information: four variables on the type of vegetation; and lastly ($4$) labels indicating the presence of pyroCb. Table \ref{intext_variabletable} includes a description of these variables and their shorthand. All variables are available for a $200 \times 200$ grid of $1$\,km$^2$ pixels. For this study we complemented the \textsc{Pyrocast} database with the altitude from the world elevation model \citep{etopo5} (denoted \textit{alt}). We also calculated the magnitude of the wind at $10$\,m and at $250$\,hPa (denoted \textit{uv10} and \textit{uv250}, respectively) using the relevant \textit{u} (east-west) and \textit{v} (north-south) components. The six geostationary channels and $22$ meteorological and fuel variables are defined in Table \ref{intext_variabletable}.



\begin{table}[ht]
\small
\caption{Geostationary and ERA5 fuel and meteorological variables. Description and motivation for selecting as a candidate cause of pyroCb.}
\label{intext_variabletable}
\centering
\begin{tabular}{p{.15\linewidth} p{.35\linewidth} p{.4\linewidth}}
\toprule
Variable & Description & Sensitive to \\
\midrule
\textit{ch}$1$ & $\phantom{\{0}0.47$\,\textmu m & smoke, haze\\
\textit{ch}$2$& $\phantom{\{0}0.64$\,\textmu m & terrain type\\
\textit{ch}$3$ & $\phantom{\{0}0.86$\,\textmu m & vegetation\\
\textit{ch}$4$ & $\phantom{\{0}3.9$\,\textmu m & thermal emissions \& cloud ice crystal size \\
\textit{ch}\{\textit{5},\textit{6}\} & $\{11.2,13.3\}$\,\textmu m & thermal emissions \& cloud opacity \\
\midrule
\{\textit{u},\textit{v}\} & \{\textit{u},\textit{v}\} comp. of wind at $250$\,hPa & upper-level dynamics which influence rising motion \\
\{\textit{u},\textit{v}\}$10$ & $10$\,m \{\textit{u},\textit{v}\} component of wind & change in fire intensity and spread \\
\textit{fg}$10$ & $10$\,m  gusts since prev. post-processing & (same as above)\\
\midrule
\textit{blh} & boundary layer height & height of turbulent air at the surface \\
\midrule
\textit{cape} & convective available potential energy & energy for air to ascend into atmosphere\\
\textit{cin} & convective inhibition & energy that will prevent air from rising\\
\textit{z} & geopotential & energy needed for air to ascend into atmosphere as a function of altitude\\
\midrule
\{\textit{slhf}, \textit{sshf}\} & surface \{\text{latent}, \text{sensible}\} heat flux & heat released or absorbed \{\text{from}, \text{neglecting}\} phase changes\\
\midrule
\textit{w} & surface vertical velocity & ascent speed of the plume from the wildfire\\
\midrule
\textit{cv}\{\textit{h},\textit{l}\} & fraction of \{\text{high}, \text{low}\} vegetation & available fuel for the wildfire \\
\textit{type}\{\textit{H},\textit{L}\} & type of \{high, low\} vegetation & (same as above)  \\
\midrule
\textit{r}\{$650$,$750$,$850$\} & rel. humidity at \{$650$,$750$,$850$\}\,hPa & condensation of vapour into clouds \\
\bottomrule
\end{tabular}
\end{table}

\section{Invariant Causal Prediction}
\label{ICP}

\subsection{Problem Statement}
Consider the setting of independent and identically distributed samples of the random vectors $X=(X_1, X_2,...,X_p)^\top \in \Real^p$, $E=(E_1,E_2,...,E_q)^\top \in \Real^q$, and $Y \in \{0,1\}$. For a variable of interest $Y$, $E$ is a set of environment variables which may be causes of $X$ but neither direct causes of $Y$, nor effects (direct or indirect) of $Y$.  Let us assume that $Y$ is generated \emph{causally} from a subset $S^* \subseteq \{1,...,p\}$ of the $p$ variables considered, hence there is causal sufficiency, and $Y$ is generated from a Structural Causal Model obeying:
\begin{align}
    Y = g(X_{S^*}, \epsilon),\phantom{bl} \epsilon \sim F,\phantom{bl} \epsilon \indep X_{S^*},   
\end{align}
where $g$ and $F$ are arbitrary function and distribution, respectively.  A set $S \subseteq \{1,...,p\}$ is a generic subset of the full set of candidate causes. We refer to $S$ and $X_S$ interchangeably for the sake of brevity. The problem we deal with in this work is to infer the set $S^*$ of direct causes of the variable of interest $Y$. 

\subsection{Linear and Nonlinear ICP}

The Invariance Causal Prediction (ICP) framework \citep{peters2016} is based on the observation that $Y$ is independent of $E$ given $X_{S^*}$, denoted $Y \indep E | X_{S^*}$. Assuming we have a set of candidate causes $X$ that includes $S^*$, the causal subset, and an environment variable $E$ that we know does not directly cause $Y$ or is an effect of $Y$, then we can search for $S^*$ by applying a conditional independence test $Y \indep E|X_S$ on $Y$, $E$ and subsets $S \subseteq \{1,...,p\}$ of $X$. ICP then selects as causal variables the intersection of all those subsets $S$ where the corresponding conditional independence test is not rejected:
\begin{align}
    \hat{S}^* = \bigcap\limits_{S:p_S > \alpha} S.   
\end{align}
Here $p_S$ is the $p$-value associated to the conditional independence test of $Y \indep E|X_S$, with the null-hypothesis corresponding to conditional independence. We do not reject conditional independence, at significance level $\alpha$, if $p_S>\alpha$.

In \citep{peters2016} an algorithm to implement ICP is presented that assumes linear relationships between causes $X$ and effects $Y$, and a categorical, univariate variable $E$. In \citep{Heinze2018} a more general  algorithm is introduced that allows for nonlinear relationships between cause and effects and for continuous and multivariate environment variables $E$. This algorithm uses a conditional independence test based on regression models of $Y$ on $X$ and $Y$ on $(X,E)$.

However, there are two problems with implementing these algorithms:
\begin{itemize}
    \item Unavailability of a conditional independence test for a binary variable $Y$ and continuous, multivariate $X$ and $E$.
    \item Combinatorial nature of ICP: it is not feasible to realize all tests needed to include all sets in the case of a large powerset (with cardinality over $250$ million). 
\end{itemize}
These issues are addressed with greedy-ICP.

\subsection{Greedy-ICP}

As in \citep{Heinze2018} we use an \emph{invariant target prediction} conditional independence test. This consists of predicting the target $Y$ with two models. The first uses $X_S$ only as
predictors, while the second uses both $X_S$ and $E$.  If the null-hypothesis is true, and $Y \indep E|X_S$, then the out-of-sample performance of both models is statistically indistinguishable. To deal with the binary nature of our target variable $Y$ we compare the receiver-operator curves (ROC) of both models by performing the non-parametric hypothesis test on the difference of the area under the curve (AUC) introduced by \citep{DeLong1988}. To address with the large amount of candidate causal variables we use the following \emph{greedy-ICP} search algorithm to obtain a causal importance ordering of the variables:

The algorithm is initialised by setting $S:=\{1,...,p\}$ to be all candidate causal variables. While the cardinality of $S$ is larger than one: 
\begin{enumerate}
    \item For each variable $i$ in $S$ perform conditional independence test $Y \indep E | X_{S'}$ where $S'=S \setminus \{i\}$,  obtaining corresponding $p$-value $p_i$.
    \item Remove variable $j$ from $S$ such that $j = \argmax_i p_i$ and redefine $S \leftarrow S \setminus \{j\}$.
\end{enumerate}

The causal predictors, at a significance level $\alpha$, are taken to be all those not eliminated when the first $p$-value below $\alpha$ is computed. It is not guaranteed that the set inferred by greedy-ICP is a superset of that inferred by the exhaustive ICP. Its use in this work is as an exploratory tool to obtain a causal importance ordering of the variables and as a tool to obtain a smaller set of candidate causes on which we can perform the full exhaustive ICP. 

\section{Setup for PyroCb}
\label{setup}

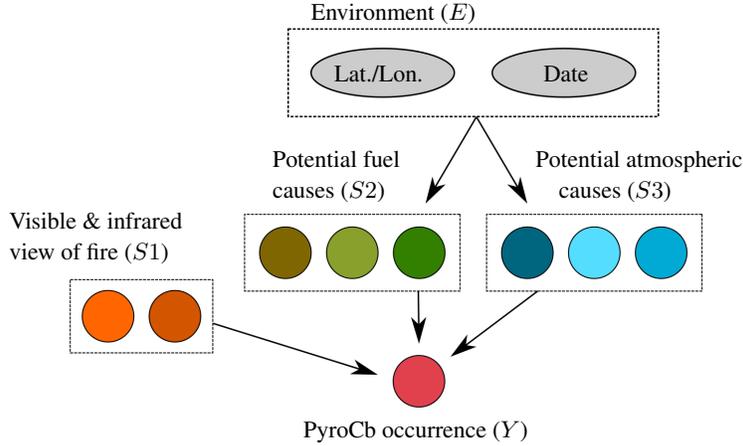
\begin{figure}[htbp]
    \centering
    \begin{overpic}{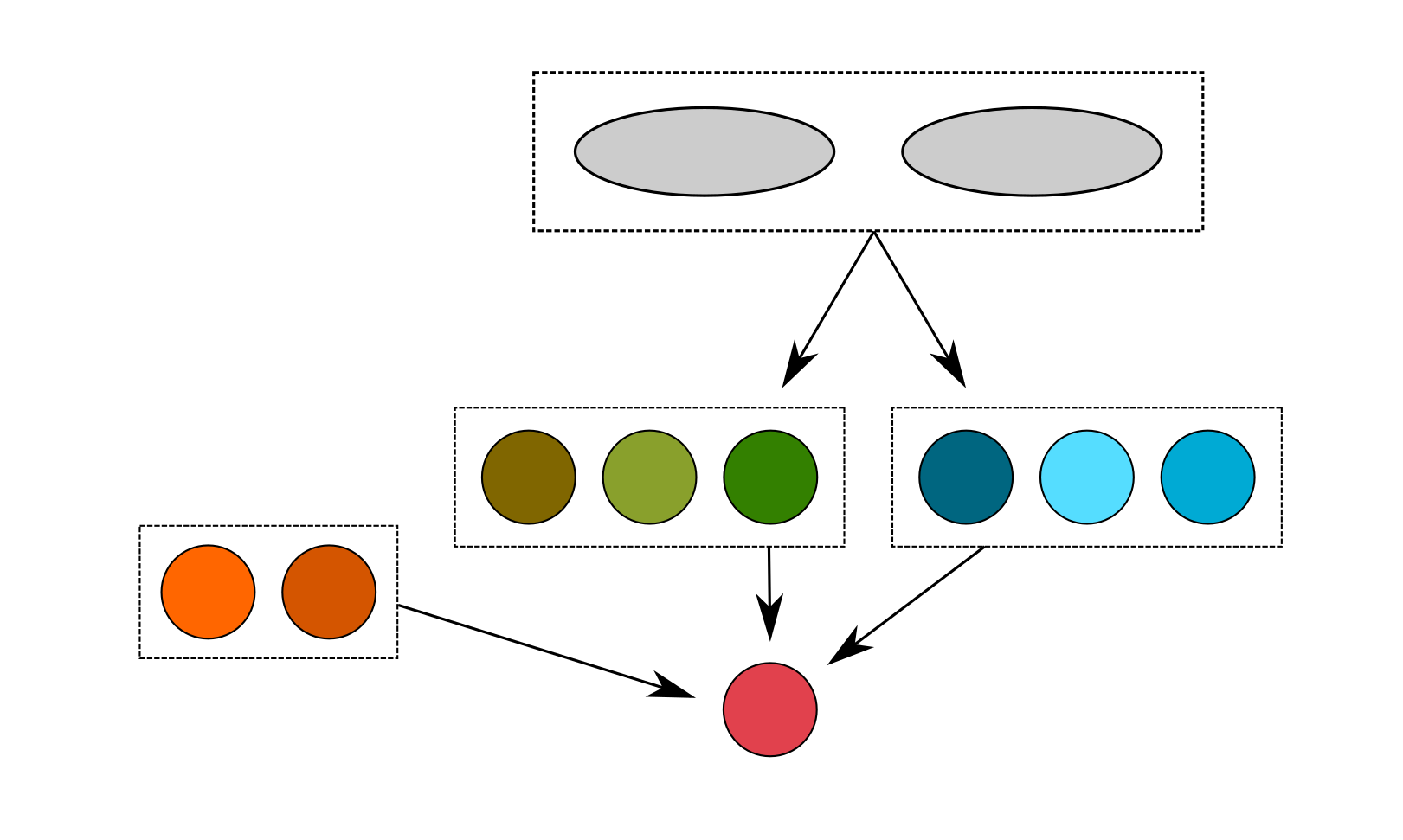}
        \put(41, 56){\small Environment ($E$)}
    
        \put(44, 48){\small Lat./Lon.}
        \put(71, 48){\small Date}
    
        \put(36, 37){\small Potential fuel}
        \put(36, 33){\small causes ($S2$)}
    
        \put(70, 37){\small Potential atmospheric}
        \put(73, 33){\small causes ($S3$)}
    
        \put(2, 29){\small Visible \& infrared}
        \put(2, 25){\small view of fire ($S1$)}
    
        \put(40, 2){\small PyroCb occurrence ($Y$)}
    \end{overpic}
    \caption{Schematic representation of assumptions made about causes of pyroCb. Boxes group variables (circles) into categories relevant to the pyroCb study. Arrows indicate assumed causal relationships between variables. Arrows that start/end at a given box indicate that at least one of its variables is assumed to be a cause/effect with respect to the end/start point of the arrow. The causal task of ICP is to use that  $Y \indep E|X_{S^*}$ (assuming $S^*$ is the set of true causes), in order to find $S^*$, a subset of biophysical ($S1$ \& $S2$) and atmospheric descriptors ($S3$) such the prediction model for $Y$ based on $X_{S^*}$ is stable across environments.}
    \label{conceptualICP}
\end{figure}

To determine the causes of pyroCb formation, the geostationary imagery and the $22$ meteorological and fuel variables are used as our candidate causes $X$. Figure \ref{conceptualICP} summarizes the ICP framework for inferring the causes of pyroCb.

Each observation consists of a $200 \times 200$ pixel grid of observations. With the exemption of the discrete categorical variables, we aggregate these observations by taking $11$ statistics: mean, standard deviation, minimum, maximum, and seven quantiles corresponding to the $1$, $5$, $25$, $50$, $75$, $95$ and $99$ percentiles. We arrange these predictors in a new vector  $X \in \Real^{296}$. However, note that when applying ICP we refer to the original $28$ vector valued variables meaning we include or exclude all the different spatial aggregations of a given variable in a model jointly. As environment variables we used the latitude, longitude, and start Julian date of the wildfire events.

The conditional independence test used involves fitting models of $X_S$ and $E$ on $Y$ for different subsets $S$. We use Random Forest (RF) models consisting of $100$ trees, where tree splits were chosen to minimise the Gini impurity. Trees were grown to a maximum depth of $10$ splits. Each tree was trained on a bootstrap subsample of the data. In evaluating the quality of splits, weighting was used in order to compensate for unbalanced labels.

The hypothesis test, introduced in \citep{DeLong1988}, on the difference of the AUC between the model containing the environment and that excluding the environment is used as a conditional independence test. If $E$ is conditionally independent of  $Y$ given a subset $S$ then including $E$ should not result in a statistically distinguishable AUC. The test is carried out on out-of-sample predictions of $Y$. For this a $5$-fold cross validation (CV) scheme is implemented where each fold is constructed at a wildfire event level by randomly assigning each event to one of five clusters. The generalization performance obtained with this CV scheme is denoted \emph{event generalization} performance.  

We validate results from the greedy-ICP algorithm by considering \emph{spatial generalization}. For this we used a $5$-fold cross validation scheme where each fold is constructed at a wildfire event level by applying a $k$-means algorithm on the latitude and longitude of all wildfire events. Figure \ref{spatialClustering} in the \nameref{appendix} delineates these clusters. 

\section{Results}
\label{section:results}
\begin{figure}[htbp]
    \centering
    \begin{subfigure}{.49\textwidth}
        \includegraphics[width=\textwidth]{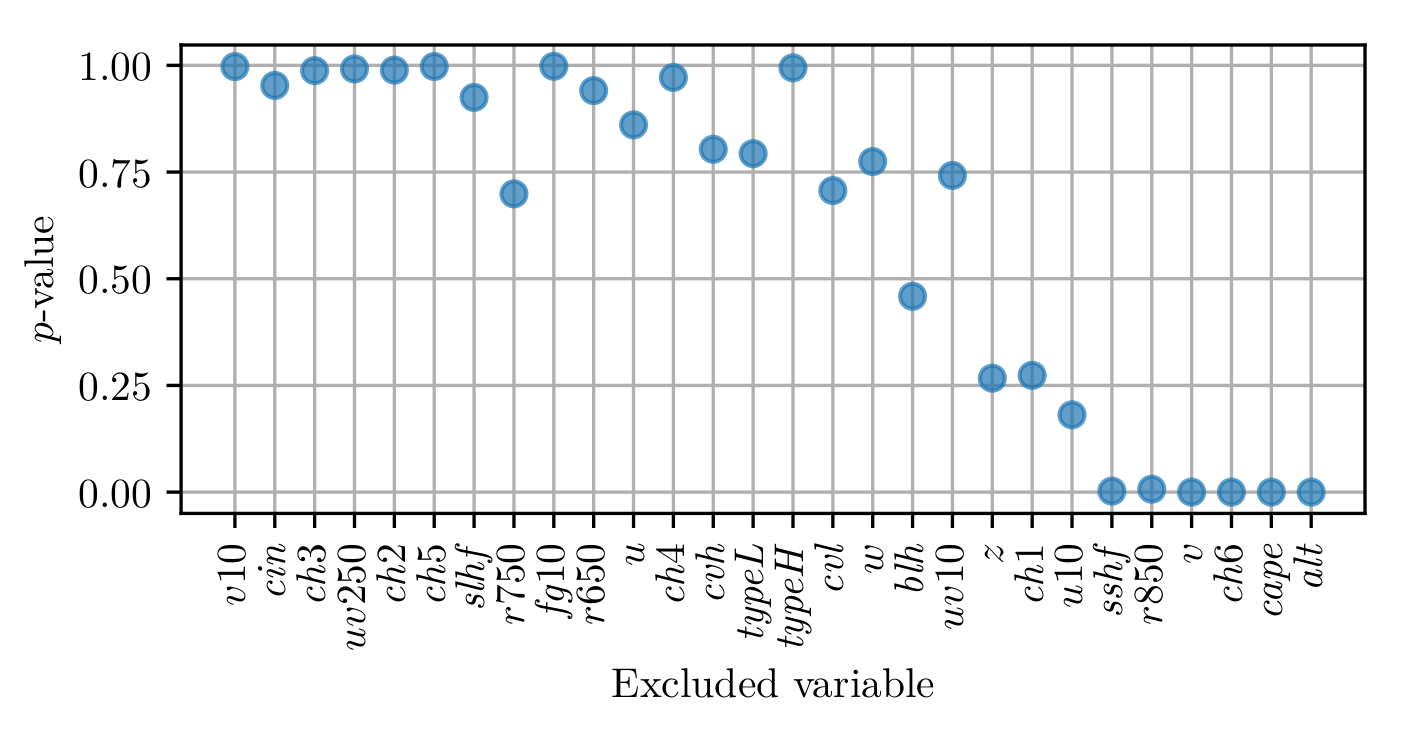}
    \end{subfigure}
    \hfill
    \begin{subfigure}{.49\textwidth}
        \includegraphics[width=\textwidth]{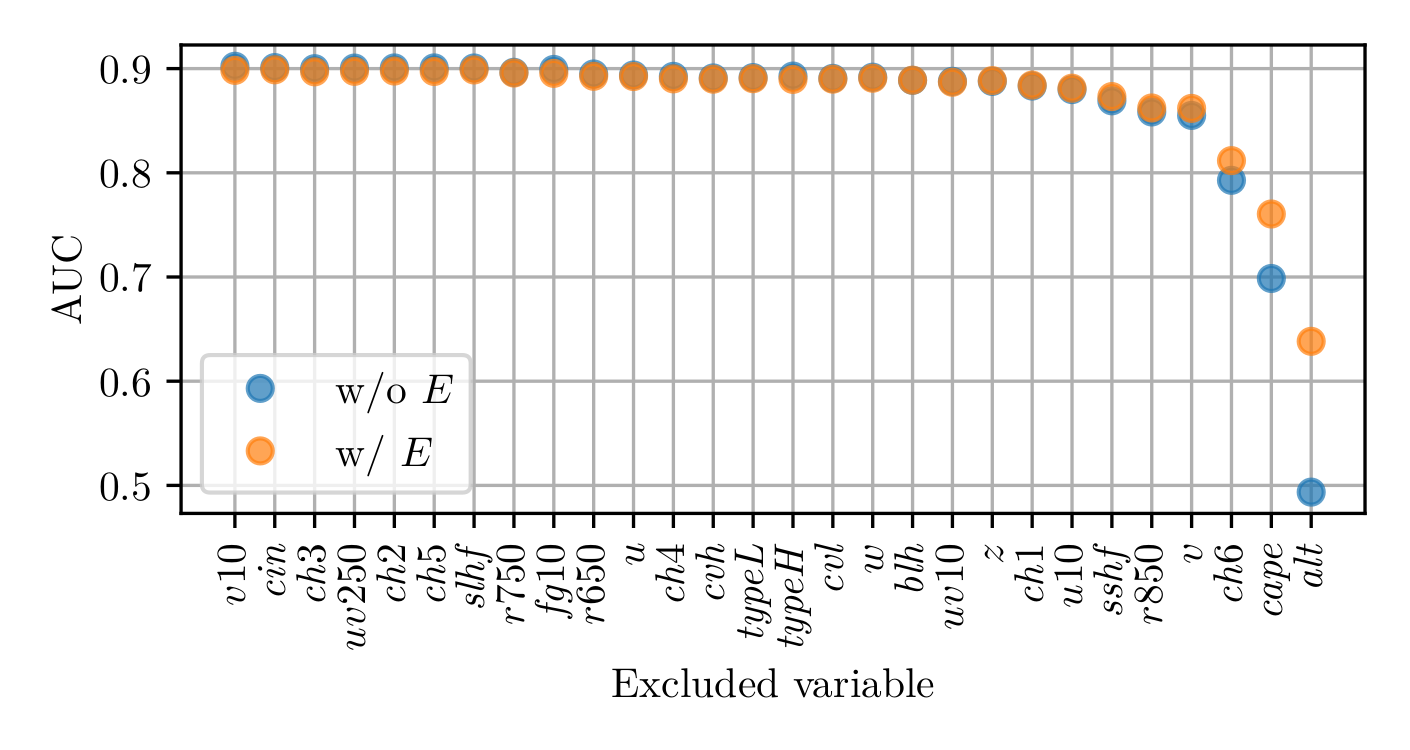}
    \end{subfigure}
  \centering
  \caption{Sequence of $p$-values (left) and AUC values (right) for sequence of candidate cause sets produced by greedy ICP. In both cases, the abscissa, read from left to right, shows the order in which causal candidates are excluded from consideration by the greedy ICP algorithm. On the left, for every variable $i$, the $p$-value corresponding to the null-hypothesis that $Y \indep E | X_S$, where $S$ is composed of excluding all variables up to variable $i$, is shown. Generally, as more variables are excluded the causally necessary hypothesis of independence starts to be rejected. On the right we show the AUC, for prediction models including the environment variables (w/ $E$, in blue) and excluding these variables (w/o $E$ in orange), for the same shrinking sequence of candidate causal predictors. As more variables are excluded, including environment information increasingly helps prediction models.)}
  \label{fig:greedyICP}
\end{figure}

Figure \ref{fig:greedyICP} shows the sequence of excluded variables, the corresponding maximum $p$-value of the conditional independence tests and the corresponding AUC of the model with and without the environment $E$. 
The order of exclusion of variables with greedy-ICP is the following: \textit{v}$10$, \textit{cin}, \textit{ch}$3$, \textit{uv}$250$, \textit{ch}$2$, \textit{ch}$5$, \textit{slhf}, \textit{r}$750$, \textit{fg}$10$, \textit{r}$650$, \textit{u}, \textit{ch}$4$, \textit{cvh}, \textit{typeL}, \textit{typeH}, \textit{cvl}, \textit{w}, \textit{blh}, \textit{uv}$10$, \textit{z}, \textit{ch}$1$, \textit{u}$10$, \textit{sshf}, \textit{r}$850$, \textit{v}, \textit{ch}$6$, \textit{cape} and \textit{alt}. Notice that the final variable, alt, has no $p$-value $p_i$ as this refers to the the set $S \setminus \{i\}$ and excluding it would leave the set empty. We can see that with the greedy search algorithm we would exclude all variables except $sshf$, \textit{r}$850$, $v$, \textit{ch}$6$, \textit{cape} and \textit{alt} at a $0.05$ significance level.  This is largely in agreement with what the literature has suggested in the sense that a very large and intense fire (\textit{ch}$6$, the $13.3$\,\textmu m channel sensitive to thermal emission)  mid-tropospheric moisture source (\textit{r}$850$), conditions favorable for thunderstorms (\textit{cape}), an unstable boundary layer (\textit{sshf}) are conditions necessary for pyroCb generation \citep{peterson2018model,tory2018thermodynamics,tory2021pyrocumulonimbus}. Additionally, the \textit{v} wind component at $250$\,hPa and the altitude of the location of the wildfire are found to be important causal factors. This is plausible since wind at high altitudes contributes to an unstable atmosphere and a high surface altitude will diminish the amount of energy needed for the hot moist air to rise and condense at high altitudes. 
Next we obtain independent validation of this result by performing cross-validation of random forests regression models of $Y$ on $X_S$ which follow the same sequence of excluding variables one-at-a-time as the greedy ICP algorithm. We use the random (event generalization), and  spatial (spatial generalization), cross-validation schemes described in Section~\ref{setup}. Figure \ref{spatialClustering} shows what events are included in each fold. If greedy ICP has ordered variables correctly, we expect spatial generalization performance to be more robust to excluding non-causal variables: these variables are robust to environment.

\begin{figure}[htbp]
    \centering
    \begin{subfigure}{.49\textwidth}
        \includegraphics[width=\textwidth]{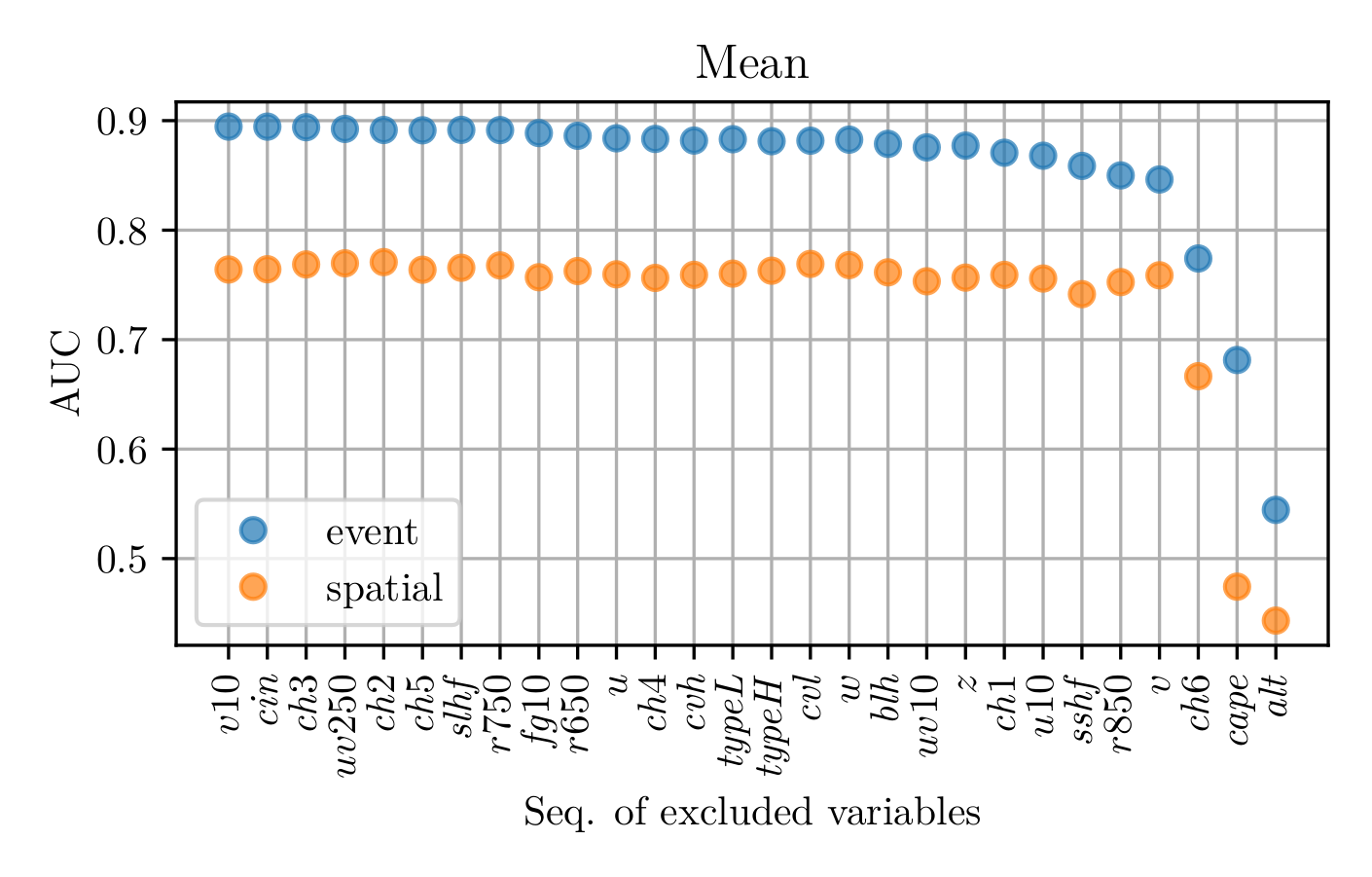}
    \end{subfigure}
    \hfill
    \begin{subfigure}{.49\textwidth}
        \includegraphics[width=\textwidth]{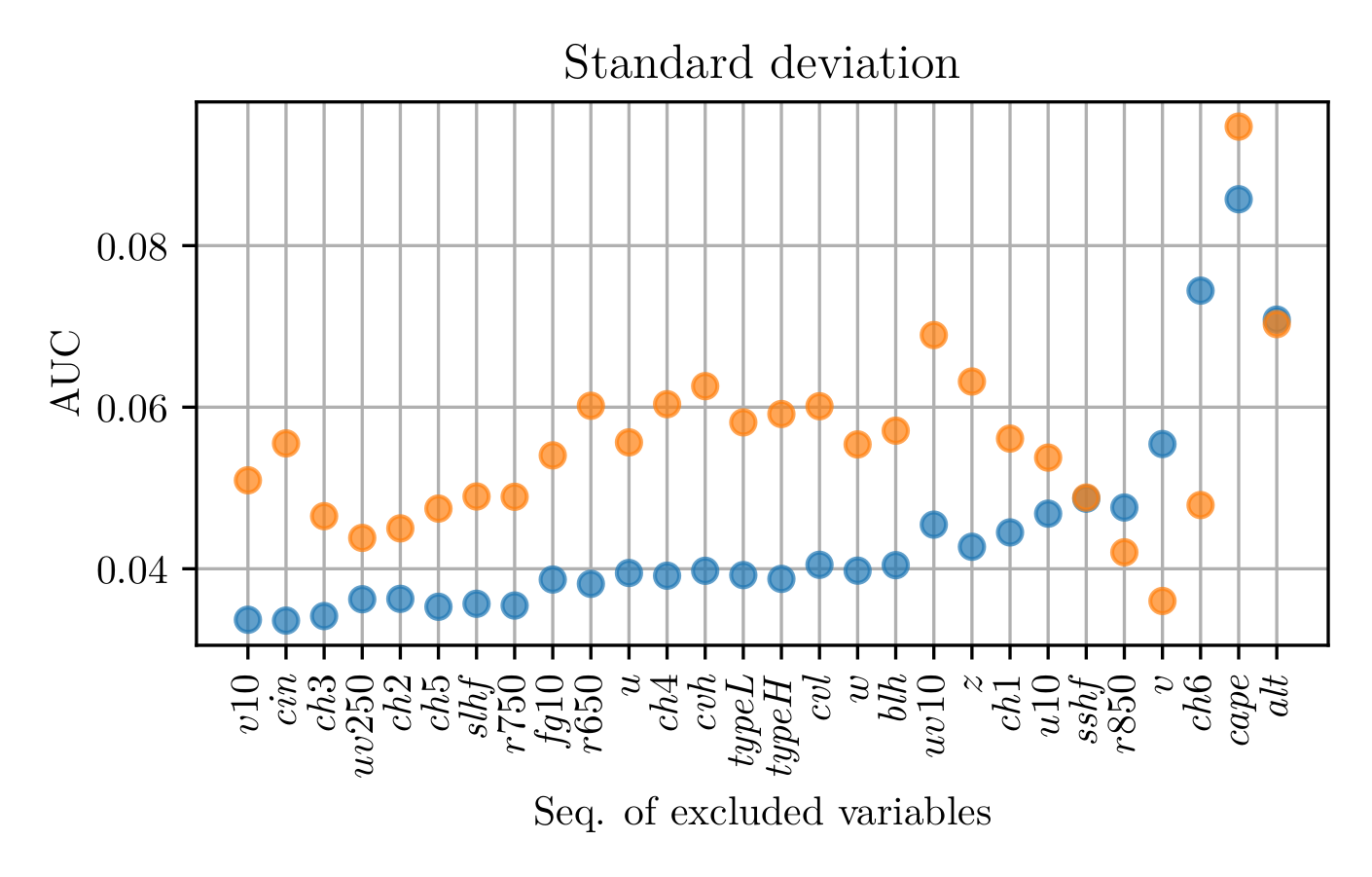}
    \end{subfigure}
  \centering
  \centering
  \caption{Validation of greedy ICP results. Since ICP should help find variables which have an invariant relationship to the target $Y$, across different environments, we use spatial-CV to verify that, as causal candidates are excluded from consideration, the spatial generalization performance remains stable, at least for those subsets of variables where the causally necessary conditional independence criterion $Y \indep E|X_S$ is not rejected. We compare to event generalization which does not measure robustness across environments.  Mean (left) and standard deviation (right) of out-of-sample AUC for event-CV (blue) and spatial-CV (orange) are shown.}
  \label{validationICP}
\end{figure}

Figure \ref{validationICP} indeed shows that excluding non-causal variables leads to negligible loss in spatial generalization performance as measured by mean AUC and actually improves when considering the standard deviation of AUC across spatial folds. In contrast, excluding variables deteriorates event generalization (both in mean and standard deviation) since these non-causal variables were detecting local associations between the non-causal variables and pyroCb occurrence.  

We subsequently take one of the accepted subsets found with greedy-ICP and perform a more exhaustive ICP on this subset. Although the seven-variable subset \{\textit{u}$10$, \textit{sshf}, \textit{r}$850$, \textit{v}, \textit{ch}$6$, \textit{cape}, \textit{alt}\} was accepted, we take a larger 11-variable subset  \{\textit{blh}, \textit{v}$10$, \textit{z}, \textit{ch}$1$, \textit{u}$10$, \textit{sshf}, \textit{r}$850$, \textit{v}, \textit{ch}$6$, \textit{cape}, \textit{alt}\} since excluding \textit{blh} from $S$ was the first time the $p$-value dropped below $0.5$. We consider all subsets of size eight or larger of this set. This amounts to $232$ subsets. Performing the conditional independence test $Y \indep E | X_S$ on these sets resulted in $60$ being accepted at significance level of $0.05$. Unfortunately, the intersection of the $60$ subsets is the empty set, meaning this more exhaustive version of ICP does not shed light on the causal drivers. As is pointed out in \citep{Heinze2018} this is often the case when the candidate causal predictors have high dependence among them. One solution pointed out in that work is to consider \emph{defining sets} which are sets of sets where the interpretation is that one of the sets is the true set of causal drivers but we cannot say which. However computing the defining sets for the $60$ accepted sets results in a defining set of $45$ sets of potential causal drivers. This does not get us much closer to identifying the most plausible causal variables. In Section~\ref{HSIC} of the Appendix, we deal with dependence and redundancy in our set of $11$ predictors by clustering variables together using a measure of the dependence between them and present results for exhaustive ICP on the clusters of variables.

\section{Conclusions}
Using a novel conditional independence test for $Y \indep E| X_S$
and a greedy-ICP search algorithm we identified a set of candidate causes for pyroCb: surface sensible heat flux (\textit{sshf}), relative humidity at $850$\,hPa (\textit{r}$850$), a component of wind at $250$\,hPa (\textit{v}), $13.3$\,\textmu m thermal emissions (\textit{ch}$6$), convective available potential energy (\textit{cape}), and altitude (\textit{alt}). This set is plausible when contrasted with the existing literature. There are other sets of predictors that are causally viable in the sense that they obey the causally necessary conditional independence condition and  are similarly minimal (contain no additional non-causal information). As more exhaustive implementations of ICP suggest, these sets have little overlap with the one identified with greedy-ICP. This is likely because, as with most real-world datasets, there is a lot of dependence and redundancy in the dataset. More exhaustive implementation of the original ICP algorithm either yielded the empty set, or a seemingly insufficient set of predictors to more satisfyingly describe the causes of pyroCb generation. Future work will involve learning  invariant causal representations of the data such that pyroCb occurrence is independent of the environment given these representations, they are minimal in the sense that they exclude non causal information, and such that they allow us to circumvent the combinatorial nature of ICP (large number of hypothesis tests needed) and its problems dealing with dependent variables. 

\newpage
\section*{Acknowledgements}
 This work is the result of the 2022 Frontier Development Lab Europe Aerosols challenge. We are grateful for the support of the organizers, mentors, and sponsors. In particular, we would like to thank David Peterson, Michael Fromm, Annastasia Sienko, and Raul Ramos for their insight and help.
 Funding for this study was also provided by the European Research Council (ERC) Synergy Grant \textit{Understanding and Modelling the Earth System with Machine Learning} (USMILE) under the Horizon 2020 research and innovation programme (Grant agreement No. 855187).

\section*{References}
\bibliography{main.bib}


\newpage
\appendix
\section{Appendix}
\label{appendix}

\subsection{\textsc{Pyrocast} Database}
The following description of the database is reproduced from the companion \textsc{Pyrocast} paper by the same authors \citep{pyrocast}.

To create the \textsc{Pyrocast} database, historical pyroCb events were manually collated from blogs and sparse databases including the Australian PyroCb registry \citep{aus_pyrocb}, the CIMSS PyroCb Blog \citep{cimss_pyrocb}, Annastasia Sienko's master's thesis \citep{sienko}, and the PyroCb Online Forum \citep{pyrocb_forum}. The events were then matched to known historical wildfire events in the GlobFire database \citep{artes2019global} to determine the start and end date of each fire. PyroCb events that could not be associated with any wildfire were given arbitrary dates three days before and after the pyroCb sighting. 

Geostationary satellite imagery from Himawari-8, GOES16, and GOES17 was then matched to the wildfire locations and dates. The spatial coverage of the three satellites overlaps with most of the recorded pyroCb located in North America and Australia. The satellite imaging instruments also have a high temporal ($10$ min) and spatial resolution ($0.5$-$2$\,km) to study the evolution of the wildfires. Six wavelength channels ($0.47$\,\textmu m, $0.64$\,\textmu m, $0.86$\,\textmu m, $3.9$\,\textmu m, $11.2$\,\textmu m, and $13.3$\,\textmu m), based on their sensitivity to aerosol, vegetation type and temperature at different altitudes, were chosen to best detect and predict pyroCb. Images were downloaded on an hourly basis during local daytime hours for the wildfire locations from Amazon Web Services \citep{aws_goes, aws_himawari} using a custom parallelisation pipeline. All the images were interpolated to $1$\,km resolution and cropped to $200 \times 200$ pixel images.

The cropped geostationary images were then fed into a pyroCb detection algorithm developed by \citet{peterson2017detection} at the US Naval Research Laboratory (NRL). This algorithm uses relative differences in brightness temperature between the $3.9$\,\textmu m, $11.2$\,\textmu m and $13.3$\,\textmu m geostationary images to label individual spatial pixels as either containing or not containing a pyroCb cloud. Finally, the geostationary images were matched with meteorological and fuel data from a climate reanalysis model. For this study, ERA5 reanalysis \citep{hersbach2020era5} was used as it is global, up-to-date, accurate, has a high temporal resolution (hourly), and is easily accessible via the Copernicus Data Store API \citep{cdsapi}. Nineteen variables were downloaded and interpolated to the geostationary grid.

To the best of the authors knowledge, the \textsc{Pyrocast} database is the most comprehensive global pyroCb database. It includes over $148$ pyroCb events linked to $111$ wildfires between 2018 and 2022, equivalent to over $18\,000$ hourly observations. Most importantly, it is science and machine learning ready, allowing for the systematic study of the characteristics and causes of pyroCb.

\subsection{ICP on Clusters of Variables}
\label{HSIC}

\begin{figure}[h]

  \centering
  
  \includegraphics[scale=0.5]{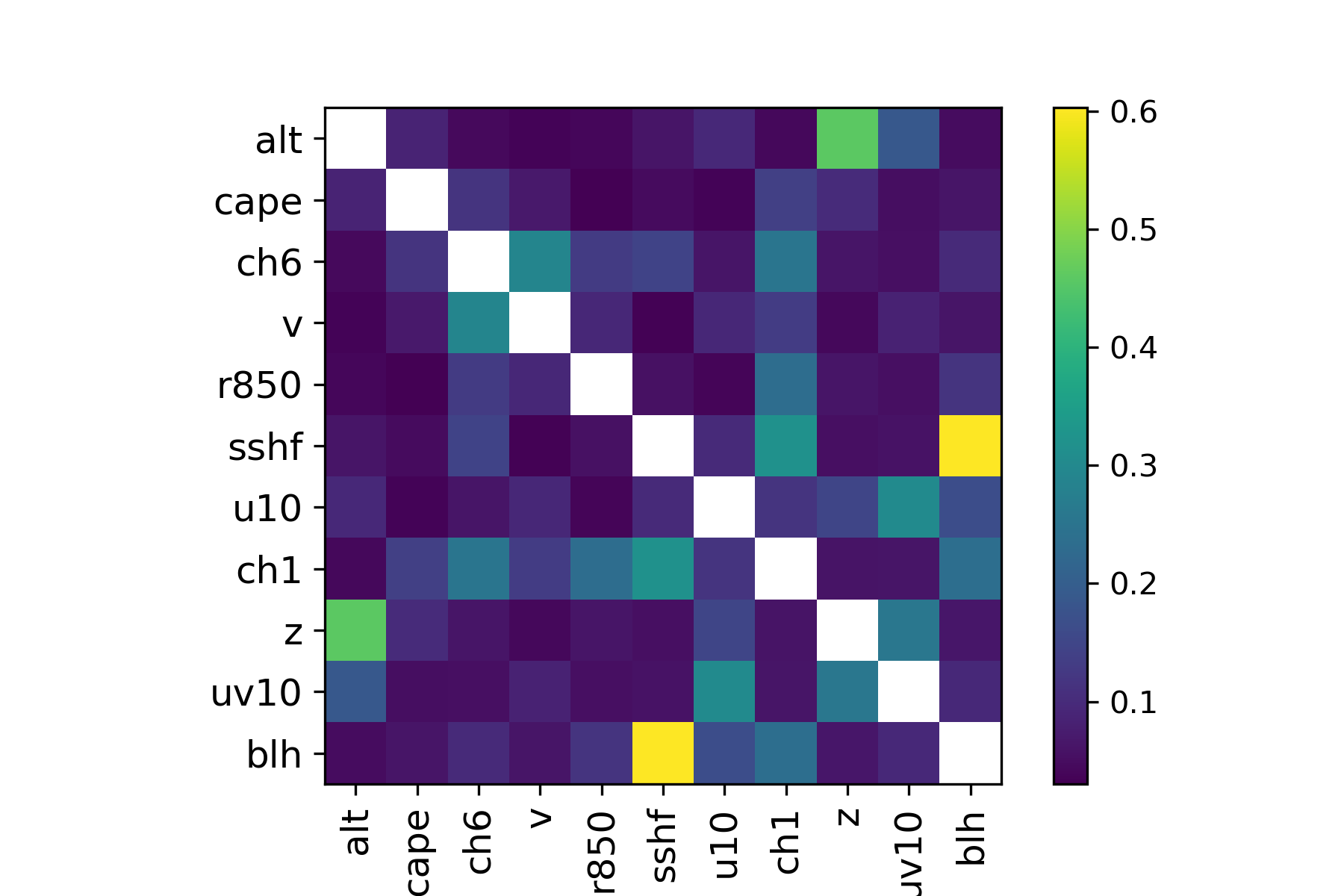}
  \caption{Normalized HSIC between each of the $11$ variables considered.}
  
  \label{hsic}
\end{figure}

In order to deal with dependence and redundancy in the set of $11$ predictors, identified in the greedy-ICP algorithm, we measure the dependence between the $11$ candidate causal variables in order to cluster variables together. Figure \ref{hsic} shows the Hilbert Schmidt Independence Criterion (HSIC) measure of dependence between the $11$ variables. 

HSIC, introduced in \citep{HSIC}, is a good choice for this case because of (1) the non-linear nature of the relationship between the predictor variables and because, (2) each variable is multivariate and includes statistical summaries such as the mean and standard deviation, of the variable across the $200 \times 200$\,km grid. We use the matrix of dependencies shown in Figure \ref{hsic} to cluster each variable with all other variables with which it has an HSIC measure of $0.25$ or larger. This results in the following non-mutually exclusive clusters:\{\textit{alt}, \textit{z}\},  \{\textit{cape}\}, \{\textit{ch}$6$, \textit{ch}$1$\}, \{\textit{v}\}, \{\textit{r}$850$\},\{\textit{sshf}, \textit{ch}$1$, \textit{blh}\}, \{\textit{uv}$10$, \textit{z}, \textit{v}$10$\},\{\textit{ch}$1$, \textit{ch}$6$, \textit{sshf}, \textit{blh}\}, \{\textit{z}, \textit{alt}, \textit{u}$10$, \textit{uv}$10$\}, \{\textit{uv}$10$, \textit{u}$10$, \textit{z}\}, and \{\textit{blh}, \textit{sshf}, \textit{ch}$1$\}. We repeat the exhaustive ICP considering this time all subsets of size eight or larger of these $11$ clusters. Since the clusters are not mutually exclusive, the resulting subsets have duplicate variables. Once duplicates are removed this results in $28$ unique subsets. We performed the conditional independence test on these sets to obtain $12$ subsets for which the conditional independence test was not rejected at a significance level of $0.05$. The intersection of these subsets is \{\textit{z}, \textit{uv10}, \textit{alt}, \textit{sshf}\}.

As before, altitude at the location of the wildfire (\textit{alt}, \textit{z}) and an unstable boundary layer (\textit{sshf}) are found to be causal drivers of pyroCb. However the variables \textit{r}$850$, \textit{v}, \textit{ch}$6$ and \textit{cape}, suggested by results from greedy-ICP are not found. It is likely that dependence and redundancy between variables is still an issue causing the exhaustive ICP method to identify sets of variables with a small overlap.

\subsection{Additional Figures}

\begin{figure}[h]
    \centering
        \includegraphics[scale=0.24]{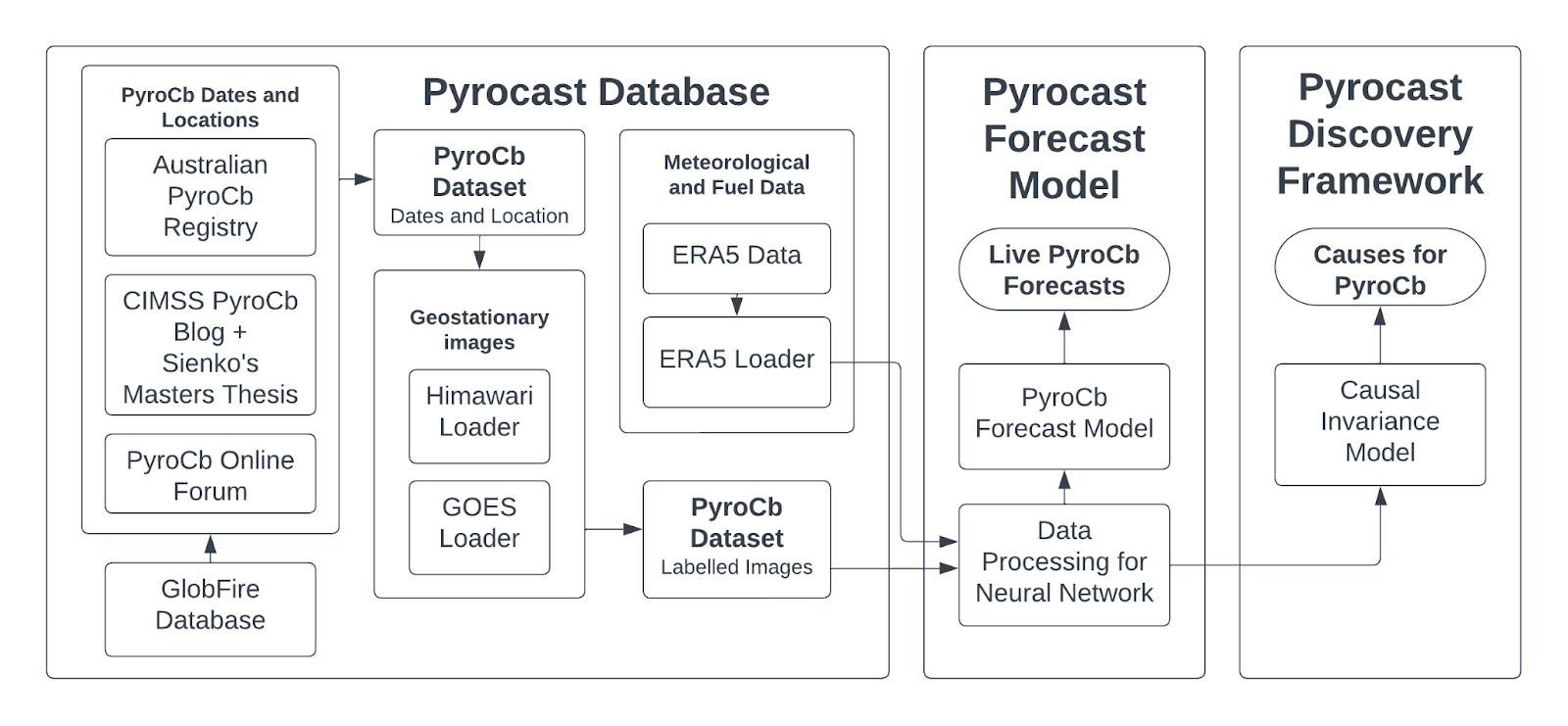}
        \caption{Diagram of \textsc{Pyrocast} pipeline. The pipeline has three components: (1) \textsc{Pyrocast} database, including pyroCb event labels, geostationary imagery in visible and infrared wavelengths, fuel descriptors and meteorological descriptors; (2)  \textsc{Pyrocast} forecast model, a 6-hour ahead machine learning forecast model for predicting pyroCb occurrence from intense wildfires, and (3)  \textsc{Pyrocast} discovery framework, with tools for understanding the causes of pyroCb events, among which, are those presented in this work. More information on the first two components can be found in \citep{pyrocast} and at \url{https://doi.org/10.56272/fpib2524}.  }
        \label{fig:pipeline}
\end{figure}

\begin{figure}[h]
    \centering
        \includegraphics[scale=0.4]{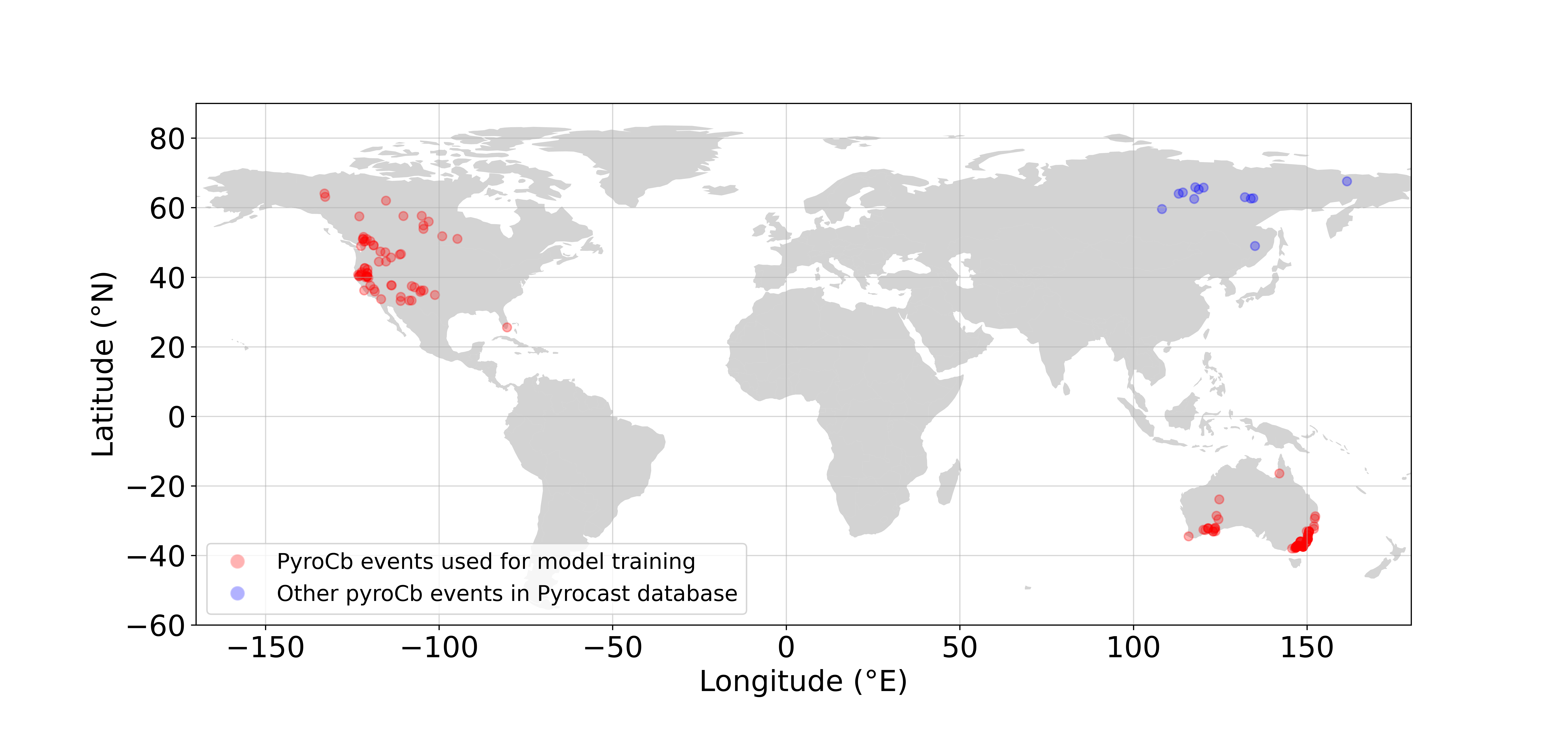}
        \caption{Spatial distribution of pyroCb events in the \textsc{Pyrocast} database.}  
        \label{fig:pyrocb_map}
\end{figure}

\begin{figure}[h]
    \centering
        \includegraphics[scale=0.4]{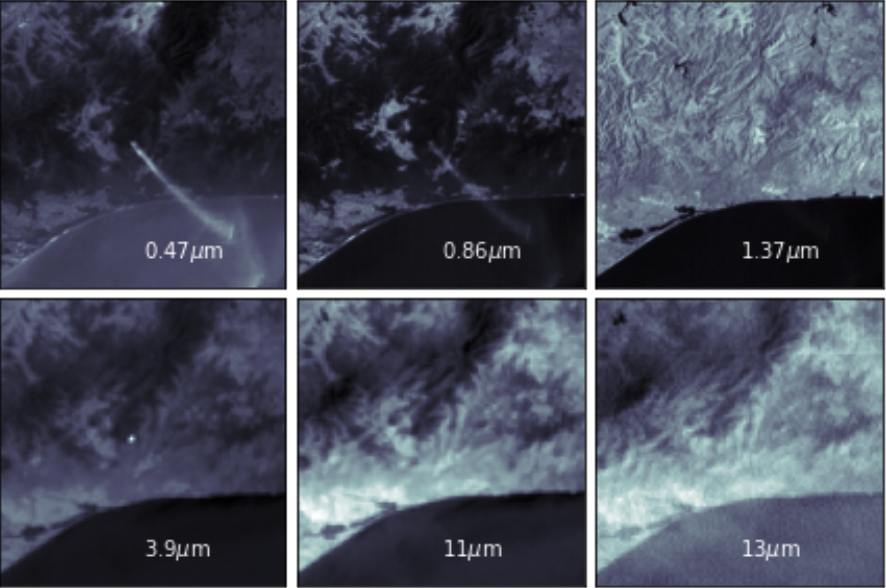}
        \caption{Example of geostationary images over the six wavelength channels of a wildfire event in Timbarra, Australia (January 2019). The centre of the wildfire can clearly be seen as a bright dot in the centre of the $3.9$\,\textmu m image, which is most sensitive to high temperatures, while the smoke plume emanating from the fire can be seen most clearly in the $0.47$\,\textmu m image, which is most sensitive to small-particle aerosols.}
        \label{fig:geo}
\end{figure}

\begin{figure}[h]
    \centering
        \includegraphics[scale=0.25]{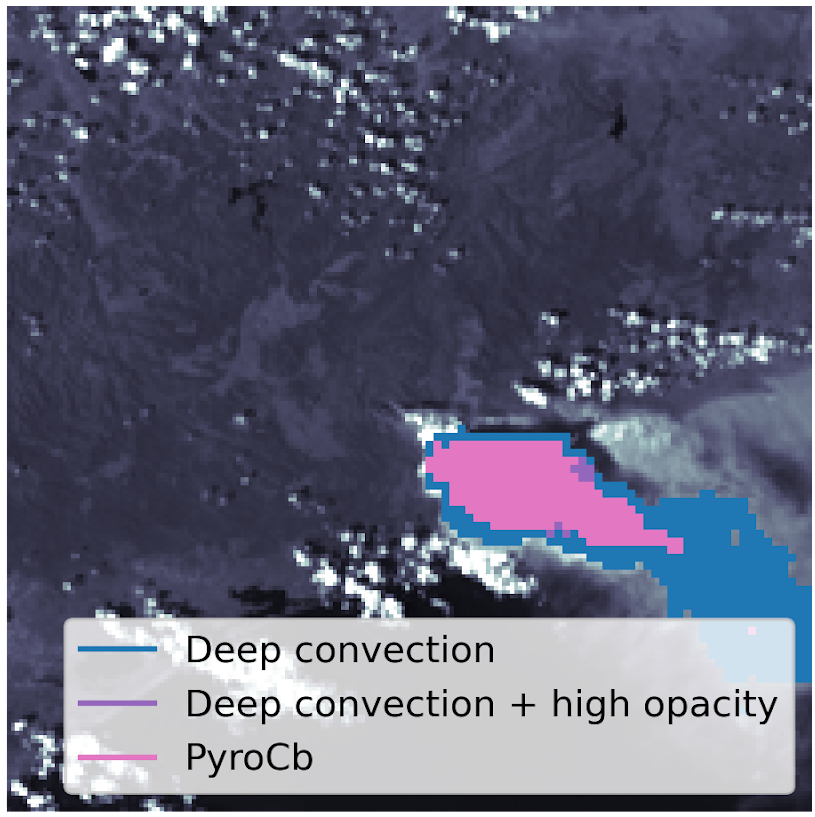}
        \caption{Example of NRL label masks overlaid onto geostationary satellite image (averaged over the $0.47$\,\textmu m, $0.64$\,\textmu m and $0.86$\,\textmu m channels) of a wildfire event in Timbarra, Australia (January 2019).}
        \label{fig:nrl}
\end{figure}

\begin{figure}[h]

  \centering
  
  \includegraphics[scale=0.8]{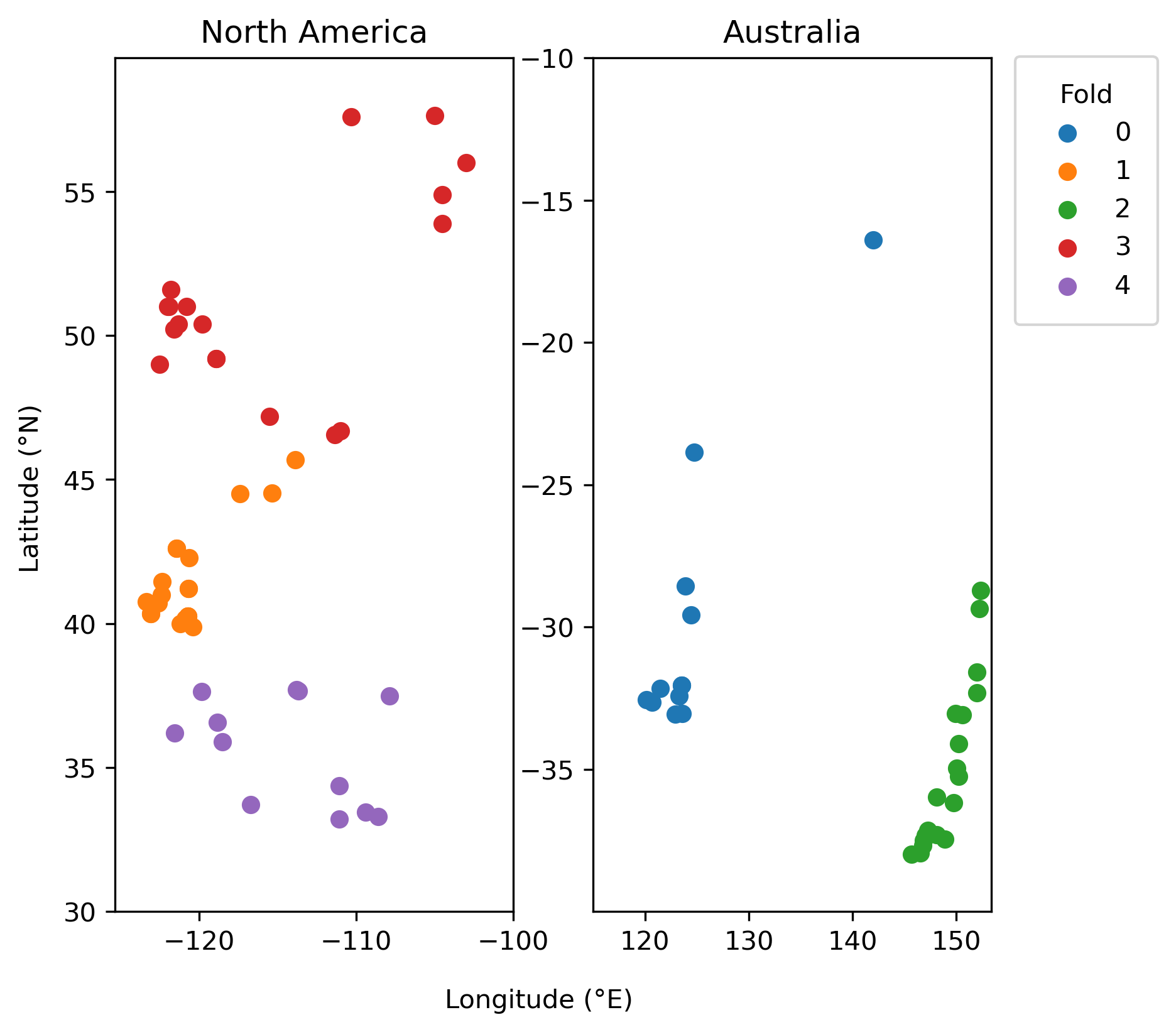}
  \caption{Clusters based on latitude and longitude for spatial cross-validation.}
  \label{spatialClustering}
\end{figure}

\end{document}